\documentclass[]{llncs}
\usepackage{graphicx}
\usepackage{makecell}
\usepackage{tabularx}
\begin{document}
%
\title{Enhancing Entity Aware Machine Translation with Multi-task Learning}

%
%
\author{An Trieu \and
Phuong Nguyen \and 
Minh Le Nguyen}

\authorrunning{Trieu et al.}
%
%

\institute{Japan Advanced Institute of Science and Technology, Ishikawa, Japan\\
\email{\{antrieu, phuongnm, nguyenml\}@jaist.ac.jp}\\
}
\maketitle              
\begin{abstract}
Entity-aware machine translation (EAMT) is a complicated task in natural language processing due to not only the shortage of translation data related to the entities needed to translate but also the complexity in the context needed to process while translating those entities. In this paper, we propose a method that applies multi-task learning to optimize the performance of the two subtasks named entity recognition and machine translation, which improves the final performance of the Entity-aware machine translation task. The result and analysis are performed on the dataset provided by the organizer of Task 2 of the SemEval 2025 competition.  
\end{abstract}

\section{Introduction}
Recently, the rapid advancement of neural machine translation (NMT) models, as exemplified by Bahdanau et al. \cite{bahdanau2014neural} and Sutskever et al. \cite{sutskever2014sequence}, has significantly boosted translation quality across various languages. Despite these considerable gains, accurately translating named entities remains an ongoing challenge. The difficulty often stems from the diverse ways in which entities are expressed across languages, compounded by insufficient data to capture their myriad forms. Unlike conventional tokens, named entities require unique translations that hinge on both the surrounding context and language-specific conventions. In human translation, a translator typically examines the context, identifies the entity type, and applies the appropriate language rules.

Motivated by this complexity, recent research has increasingly focused on integrating named entity recognition (NER) with machine translation (MT). Leveraging reliable NER outputs can help MT systems better handle specialized entities, especially those that deviate from standard usage. However, the meta format of these outputs can introduce considerable noise into the translation pipeline, potentially undermining overall performance if not carefully managed. Consequently, a robust strategy is needed to seamlessly marry NER and MT outputs while minimizing the risks associated with extraneous data.

In this work, we propose a multi-task learning framework that fine-tunes an end-to-end model for entity-aware machine translation. Our approach directly transfers information gleaned from predicting entities to the translation process, thereby minimizing noise and preserving essential contextual cues. This tight coupling between NER and MT tasks allows the model to systematically account for language-specific conventions of named entities while simultaneously improving overall translation accuracy.

Our main contribution is a method that efficiently combines the strengths of both NER and MT, fostering a synergistic relationship through multi-task learning. We evaluated our approach on the dataset provided by the organizer of Task 2 of the SemEval 2025\footnote{https://semeval.github.io/SemEval2025/tasks.html} competition. The results demonstrate an improvement in handling named entities over baseline methods, underscoring the potential of this framework in advancing the broader field of machine translation.
\section{Background}
\subsection{Entity-aware machine translation}
To enhance the performance of entity-aware translation tasks, many researchers have explored integrating the output of a Named Entity Recognition (NER) system into the Machine Translation (MT) model. Zeng et al. \cite{zeng2023extract} proposed a system that incorporates information from an external NER module into the MT model via an attention mechanism. First, entities in the source sequence are detected using an external NER model, and their corresponding target-language translations are retrieved from a pre-built dictionary. Next, the positions of these entities are represented through positional embeddings, and the entity–translation pairs are used to initialize the output sequence. This approach provides the model with detailed information about the entities and how to translate them accurately.

Similarly, Maciej et al. \cite{modrzejewski2020incorporating} also leverage an external NER model to detect named entities in the source sequence. They incorporate this information by adding boundary XML tags around each detected entity. Despite its simplicity, this method effectively draws the MT model’s attention to the tagged entities, improving translation quality.

In contrast to the above approaches, Shufang et al. \cite{xie2022end} integrate entity information directly into the attention mechanism by using multitask learning to train an end-to-end model that both predicts entities and translates the source sequence. They supply additional data about the position and type of entities to the Transformer \cite{vaswani2017attention} model and jointly optimize for translation and NER, combining both losses into the final objective function.

Following the same principle, Rikters et al. \cite{rikters-miwa-2024-entity-aware} use multitask learning to improve the translation of rare words. In their system, entities detected by an external NER model are aligned with tokens in the target sequence to discover their corresponding translations. The entities and their translations are then enclosed within XML tags containing entity-type information. These annotated source–target sentence pairs are used as training data for T5 models, enabling them to learn both entity prediction and translation simultaneously.
\subsection{T5 finetuning}
T5 (Text-to-Text Transfer Transformer) \cite{10.5555/3455716.3455856} and its multilingual variant mT5 \cite{xue2020mt5} both adopt a Transformer-based encoder–decoder architecture trained on a masked-span prediction objective. By representing every NLP task as a text-to-text problem, they naturally support multitask learning: prompts or task-specific prefixes guide the model on which task to perform, and the same sequence-to-sequence framework generates the output. Through extensive pre-training on C4 (for T5) and mC4 (for mT5), these models learn robust representations transferable to a wide array of downstream tasks. In the context of machine translation, T5 and mT5 leverage their generative design to achieve strong results, including in low-resource scenarios, while also demonstrating competitive performance on tasks such as summarization, question answering, and text classification across multiple languages.

Building on these strengths, our research focuses on integrating entity information into the input sequence to direct the model’s attention toward named entities while reducing noise and strengthening the link between entities and their correct translations. We therefore propose a method to fine-tune an end-to-end MT model capable of self-correcting entity translations. To this end, we adopt the multitask mechanism of the T5 architecture to meet our objectives.

\subsection{Introduction about SemEval 2025 Task 2}
SemEval is a series of international natural language processing (NLP) research workshops. Each year’s workshop features multiple shared tasks, where teams present and compare their computational semantic analysis systems\footnote{\url{https://semeval.github.io/}}. Task~2 in SemEval~2025 highlights Entity-Aware Machine Translation (EA-MT), focusing on translation systems that accurately handle potentially challenging named entities in source text.

The task organizers provided two main datasets: a training set derived from the Mintaka dataset~\cite{sen-etal-2022-mintaka}, and public and private test sets based on Wikidata\footnote{\url{https://www.wikidata.org/}}\cite{conia-etal-2024-towards} \cite{ea-mt-benchmark}. Each dataset entry is structured in JSON, containing fields such as an identifier, a Wikidata ID, a list of entity types, the source text, one or more target translations (each with a translated string and the mentioned entity), and the source and target locales. An example training entry is shown below:

\begin{verbatim}
"id": "Q746666_0",
"wikidata_id": "Q746666",
"entity_types": [
  "Musical work"
],
"source": "Can you sing the chorus of the folk 
song Ring a Ring o' Roses?",
"targets": [
  {
    "translation": "Puoi cantare il ritornello della canzone 
    popolare Girotondo?",
    "mention": "Girotondo"
  },
  {
    "translation": "Sai cantare il ritornello del girotondo,
    la canzone popolare?",
    "mention": "girotondo"
  }
],
"source_locale": "en",
"target_locale": "it"
\end{verbatim}
\label{fig:format_data}
\section{Methods}
With the data format in Task 2 of the SemEval, the input does not provide so much information about the entities, but the types of existing entities inside the source sequence. Therefore, our system involves two phases to completely translate an input source sequence with correct entity translation: Data preparation and fine-tuning, and the application of a multitask learning during fine-tuning on the mT5 model. \\
\textbf{Data preparation}\\
With the target-language entity mentions already available, we aligned the corresponding entities in the source language by applying two methods. First, we used the large language model Qwen/Qwen2.5-VL-72B-Instruct \cite{qwen2.5} with quantization to extract source–target alignments. To mitigate potential hallucinations, we retained only those entities identified by the model that actually appear in the source sequence. Second, we employed the AWESOME framework \cite{dou2021word} to perform token-level alignment between the source and target sequences. We then combined the outputs of both methods to construct the dataset for the subsequent fine-tuning process.\\
\textbf{Multitask learning Design}\\
For the fine-tuning process, we highlight the entities extracted earlier by enclosing them in XML boundary tags. In addition, we provide the mT5 model with a new prefix that specifies the task and the desired output format. The output for this task comprises three parts, each separated by the token ``\textless{}SEP\textgreater{}'' in the following order: (1) the result of the NER process, (2) the corresponding translation of the entities, and (3) the translation of the input sequence, where entity mentions are marked with the XML tags ``\texttt{<entity>}'' and ``\texttt{</entity>}''. We keep these tags minimal to avoid introducing excessive noise during translation. If multiple entities occur in the input sequence, they are separated by the ``\texttt{|}'' token. By placing the NER output and entity translations ahead of the full translation, we prompt the model to identify and translate entity terms first, thereby focusing more effectively on them. After generation, the NER results, entity translations, and boundary tags are removed to obtain the final translation output. An example of the fine-tuning data follows:

\begin{verbatim}
Input:
ner and translation: Who was the overall Commander of Allied 
Forces in Europe?
Desired Output:
Europe | Allied Forces <SEP> Europa | alliierten Streitkräfte <SEP> Wer 
war der Oberbefehlshaber der <entity> alliierten Streitkräfte </entity>
in <entity> Europa </entity>?
\end{verbatim}
\label{fig:finetuning_data}

\section{Experiments}
\subsection{Experimental Settings}
In the evaluation phase, we train on the data provided by the SemEval~Task~2 organizers and measure performance using BLEU~\cite{papineni2002bleu}. We partition the dataset into training, validation, and testing sets with consistent ratios across all languages, as shown in Table~\ref{tab:dataset_stats}. For fine-tuning, we employ the mT5-large model \cite{xue2020mt5} (580 Million parameters) for 50~epochs, using an initial learning rate of $5\times10^{-5}$ and a batch size of 16. All experiments are conducted on an NVIDIA A40 GPU.

\begin{table}[ht]
\centering
\begin{tabular}{|c|c|c|c|}
\hline
\textbf{Language} & \textbf{Train Size} & \textbf{Dev Size} & \textbf{Test Size} \\
\hline
English-German  & 2615 & 654 & 818 \\
\hline
English-French  & 3539 & 885 & 1107 \\
\hline
English-Italian & 2392 & 599 & 748 \\
\hline
English-Spanish & 3302 & 826 & 1032 \\
\hline
\end{tabular}
\caption{Dataset statistics for each language pair.}
\label{tab:dataset_stats}
\end{table}

\subsection{Main Results}
The results we achieved are shown in the table \ref{tab:comparison}. We compares the translation performance of three approaches (mBART, baseline, and a multi-task learning model) across four language pairs: English–German, English–French, English–Italian, and English–Spanish. The numbers represent translation quality scores (e.g., BLEU). The multi-task learning approach achieves the highest scores in most cases (en-de, en-it, and en-es), highlighting its overall effectiveness. However, for English–French, the baseline model slightly outperforms the multi-task model. Meanwhile, mBART consistently shows lower performance compared to the other two methods, indicating that either additional fine-tuning or more advanced strategies are needed for it to reach competitive levels.
\begin{table}[ht]
\centering
\begin{tabular}{|c|c|c|c|}
\hline
 & \textbf{mBART} & \textbf{mT5 large} & \textbf{Multi-task learning} \\
\hline
\textbf{English-German} & 40.79 & 46.01 & \textbf{47.69} \\
\hline
\textbf{English-French} & 42.01 & \textbf{49.61} & 48.51 \\
\hline
\textbf{English-Italian} & 35.40 & 43.49 & \textbf{48.83} \\
\hline
\textbf{English-Spanish} & 45.11 & 51.12 & \textbf{54.18} \\
\hline
\end{tabular}
\caption{Comparison of translation performance among mBART, baseline, and multi-task learning.}
\label{tab:comparison}
\end{table}

\section{Conclusion}
In this work, we introduced a multi-task learning framework for Entity-Aware Machine Translation, leveraging NER outputs to guide an mT5-based translation model. By highlighting entities and translating them in tandem with the surrounding text, our approach increases precision when handling challenging or rare terms. Experimental results on the SemEval2025 Task2 dataset confirm that coupling NER and MT tasks within a single architecture yields performance gains over both mBART and a standard mT5 baseline. Although our approach underperforms slightly in certain language pairs, particularly English–French, the overall improvements underscore the importance of integrating entity information into the translation process. Future directions include expanding coverage to additional languages and entity types, exploring alternative alignment techniques, and incorporating even larger pretrained language models for more robust handling of nuanced or low-frequency entities.

\bibliographystyle{splncs04}
\bibliography{ref}

\begin{thebibliography}{10}
\providecommand{\url}[1]{\texttt{#1}}
\providecommand{\urlprefix}{URL }
\providecommand{\doi}[1]{https://doi.org/#1}

\bibitem{bahdanau2014neural}
Bahdanau, D., Cho, K., Bengio, Y.: Neural machine translation by jointly learning to align and translate. arXiv preprint arXiv:1409.0473  (2014)

\bibitem{conia-etal-2024-towards}
Conia, S., Lee, D., Li, M., Minhas, U.F., Potdar, S., Li, Y.: Towards cross-cultural machine translation with retrieval-augmented generation from multilingual knowledge graphs. In: Al-Onaizan, Y., Bansal, M., Chen, Y.N. (eds.) Proceedings of the 2024 Conference on Empirical Methods in Natural Language Processing. pp. 16343--16360. Association for Computational Linguistics, Miami, Florida, USA (Nov 2024). \doi{10.18653/v1/2024.emnlp-main.914}, \url{https://aclanthology.org/2024.emnlp-main.914/}

\bibitem{ea-mt-benchmark}
Conia, S., Li, M., Navigli, R., Potdar, S.: {S}em{E}val-2025 task 2: Entity-aware machine translation. In: Proceedings of the 19th International Workshop on Semantic Evaluation (SemEval-2025). Association for Computational Linguistics (2025)

\bibitem{dou2021word}
Dou, Z.Y., Neubig, G.: Word alignment by fine-tuning embeddings on parallel corpora. In: Conference of the European Chapter of the Association for Computational Linguistics (EACL) (2021)

\bibitem{modrzejewski2020incorporating}
Modrzejewski, M., Exel, M., Buschbeck, B., Ha, T.L., Waibel, A.: Incorporating external annotation to improve named entity translation in nmt. In: Proceedings of the 22nd annual conference of the european association for machine translation. pp. 45--51 (2020)

\bibitem{papineni2002bleu}
Papineni, K., Roukos, S., Ward, T., Zhu, W.J.: Bleu: a method for automatic evaluation of machine translation. In: Proceedings of the 40th annual meeting of the Association for Computational Linguistics. pp. 311--318 (2002)

\bibitem{10.5555/3455716.3455856}
Raffel, C., Shazeer, N., Roberts, A., Lee, K., Narang, S., Matena, M., Zhou, Y., Li, W., Liu, P.J.: Exploring the limits of transfer learning with a unified text-to-text transformer. J. Mach. Learn. Res.  \textbf{21}(1) (Jan 2020)

\bibitem{rikters-miwa-2024-entity-aware}
Rikters, M., Miwa, M.: Entity-aware multi-task training helps rare word machine translation. In: Mahamood, S., Minh, N.L., Ippolito, D. (eds.) Proceedings of the 17th International Natural Language Generation Conference. pp. 47--54. Association for Computational Linguistics, Tokyo, Japan (Sep 2024), \url{https://aclanthology.org/2024.inlg-main.5/}

\bibitem{sen-etal-2022-mintaka}
Sen, P., Aji, A.F., Saffari, A.: Mintaka: A complex, natural, and multilingual dataset for end-to-end question answering. In: Calzolari, N., Huang, C.R., Kim, H., Pustejovsky, J., Wanner, L., Choi, K.S., Ryu, P.M., Chen, H.H., Donatelli, L., Ji, H., Kurohashi, S., Paggio, P., Xue, N., Kim, S., Hahm, Y., He, Z., Lee, T.K., Santus, E., Bond, F., Na, S.H. (eds.) Proceedings of the 29th International Conference on Computational Linguistics. pp. 1604--1619. International Committee on Computational Linguistics, Gyeongju, Republic of Korea (Oct 2022), \url{https://aclanthology.org/2022.coling-1.138/}

\bibitem{sutskever2014sequence}
Sutskever, I., Vinyals, O., Le, Q.V.: Sequence to sequence learning with neural networks. Advances in neural information processing systems  \textbf{27} (2014)

\bibitem{vaswani2017attention}
Vaswani, A., Shazeer, N., Parmar, N., Uszkoreit, J., Jones, L., Gomez, A.N., Kaiser, {\L}., Polosukhin, I.: Attention is all you need. Advances in neural information processing systems  \textbf{30} (2017)

\bibitem{xie2022end}
Xie, S., Xia, Y., Wu, L., Huang, Y., Fan, Y., Qin, T.: End-to-end entity-aware neural machine translation. Machine Learning  \textbf{111}(3),  1181--1203 (2022)

\bibitem{xue2020mt5}
Xue, L., Constant, N., Roberts, A., Kale, M., Al-Rfou, R., Siddhant, A., Barua, A., Raffel, C.: mt5: A massively multilingual pre-trained text-to-text transformer. arXiv preprint arXiv:2010.11934  (2020)

\bibitem{qwen2.5}
Yang, A., Yang, B., Zhang, B., Hui, B., Zheng, B., Yu, B., Li, C., Liu, D., Huang, F., Wei, H., Lin, H., Yang, J., Tu, J., Zhang, J., Yang, J., Yang, J., Zhou, J., Lin, J., Dang, K., Lu, K., Bao, K., Yang, K., Yu, L., Li, M., Xue, M., Zhang, P., Zhu, Q., Men, R., Lin, R., Li, T., Xia, T., Ren, X., Ren, X., Fan, Y., Su, Y., Zhang, Y., Wan, Y., Liu, Y., Cui, Z., Zhang, Z., Qiu, Z.: Qwen2.5 technical report. arXiv preprint arXiv:2412.15115  (2024)

\bibitem{zeng2023extract}
Zeng, Z., Wang, R., Leng, Y., Guo, J., Tan, X., Qin, T., Liu, T.y.: Extract and attend: Improving entity translation in neural machine translation. arXiv preprint arXiv:2306.02242  (2023)

\end{thebibliography}
\end{document}